\documentclass[submission]{eptcs}
\providecommand{\event}{ICLP 2019 -- Applications Track} 

\usepackage{underscore}           
\usepackage{wasysym}
\usepackage{algorithm}
\usepackage{algpseudocode}
\usepackage{soul}
\usepackage{tikz}
\usepackage[colorinlistoftodos]{todonotes}
\usepackage{graphicx}
\usepackage{subcaption}
\usepackage{listings}
\usepackage{hyperref}

\usepackage{amsmath,amsthm,amssymb}
\theoremstyle{definition}
\newtheorem{definition}{Definition}[section]

\title{An Implementation of a Non-monotonic Logic in an Embedded Computer for a Motor-glider}
\author{Jos\'{e} Luis Vilchis Medina \qquad Pierre Siegel \qquad Vincent Risch
\institute{Aix-Marseille Univ, Université de Toulon, CNRS, LIS\\ Marseille, France}
\email{\{joseluis.vilchismedina,pierre.siegel,vincent.risch\}@lis-lab.fr} \and Andrei Doncescu \institute{LAAS-CNRS\\Toulouse, France} \email{andrei.doncescu@laas.fr}}
\def\titlerunning{An Implementation of a Non-monotonic Logic in an Embedded Computer for a Motor-glider}
\def\authorrunning{J.L. Vilchis Medina, P. Siegel, V. Risch \& A. Doncescu}
\begin{document}
\maketitle
\begin{abstract}
	In this article we present an implementation of nonmonotonic reasoning in an embedded system. As a part of an autonomous motor-glider, it simulates piloting decisions of an airplane. A real pilot must take care not only about the information arising from the cockpit (airspeed, altitude, variometer, compass\ldots) but also from outside the cabin. Throughout a flight, a pilot is constantly in communication with the control tower to follow orders, because there is an airspace regulation to respect. In addition, if the control tower sends orders while the pilot has an emergency, he may have to violate these orders and airspace regulations to solve his problem (e.g. emergency landing). On the other hand, climate changes constantly (wind, snow, hail\ldots) and can affect the sensors. All these cases easily lead to contradictions. Switching to reasoning under uncertainty, a pilot must make decisions to carry out a flight. The objective of this implementation is to validate a nonmonotonic model which allows to solve the question of incomplete and contradictory information. We formalize the problem using default logic, a nonmonotonic logic which allows to find fixed-points in the face of contradictions. For the implementation, the Prolog language is used in an embedded computer running at 1 GHz single core with 512 Mb of RAM and 0.8 watts of energy consumption.
\end{abstract}
\section{Introduction} 
\label{sec:introduction}
In this article we present an implementation of the calculation of extensions of a default theory in an embedded computer . The practical case is about rules of piloting an airplane. This is a complex human activity in terms of management of the rules of legislation, control tower, environment, risks\ldots A pilot needs to manage these rules, plus taking into account informations from the cockpit that changes every time. A cockpit is composed of six instruments such as airspeed, altimeter, compass, variometer, turn bank and artificial horizon. All this gives the states of the airplane to the pilot. In order to tackle the possible contradictions in these changing informations, we use default logic \cite{reiter1981closed}, a nonmonotonic way of reasoning which is a manner to represent the way of human reasoning \cite{DegrandeSchaub,McCarthya,McCarthyb}. Especially, default logic gains the benefits of an interpretive semantics \cite{DelScha,konolige1987relation,Sombe}.
The practical interest of this paper is to implement the computation of extensions of a default theory with minimum requirements of energy consumption and effective time of computation.
One of the most studied logic programming language in theoretical and practical terms is Prolog \cite{kifer2018declarative,sterling1990practice}. Logic programming is a programming paradigm which is based on facts and rules describing a problem in a particular domain. For this implementation Prolog is used. Prolog uses a kind of default assumption when treating negation: for instance, if a negative literal cannot be proved to be true, then it is assumed to be false. Prolog is based on a fragment of first-order logic (FOL) where rules are in the form of clauses: $\mathit{H :- \ B_1, B_2,\cdots,B_n.}$ with $\mathit{H}$ as the head, the symbol $\mathit{:-}$ as ``if'' and $\mathit{B_1, B_2,\cdots,B_n}$ as the body. Facts are clauses without body: $\mathit{H}$\footnote{Prolog is based on \emph{Horn clauses}. A Horn clause is a clause with at most one positive literal, this is  a subset of FOL.}.
The motivation of this implementation is to have Prolog running in a microcomputer to be able to calculate extensions. Later, this can be incorporate in a mobile system such as a motor-glider to simulate the decisions of the pilot.

\subsection{Classical Logic} 
Logic is a particular way of thinking, that focus on the formal principles of inference, and hence on consequences from given axioms. Formal systems, e.g., propositional, predicate, modal\ldots logics are symbolic constructions in a particular language which allows to express different ways to deal with a conclusion \cite{russell2016artificial}.
The language of propositional logic is defined as the least set of expressions satisfying: $\top$ (true) and $\bot$ (false) are formulas, and if $A$ and $B$ are formulas, so are $\lnot A$ (not A), $A \land B$ (A and B), $A \lor B$ (A or B) and $A \rightarrow B$ ($A$ implies $B$).
A \emph{proposition} can be any sentences, e.g., ``It's a sunny day'', ``Robert can pilot his airplane''. Propositional variables are denoted by a, b, c\ldots The sentence ``It's a sunny day'' could be represented by $A$, and the other sentence ``Robert can pilot his airplane'' by $B$. It can be composed to create complex sentences, e.g., ``It's a sunny day and Robert can pilot his airplane'', resulting: $A \land B$.
First-Order Logic (FOL) or predicate logic is an extension of propositional logic that includes universal and existential quantifiers, respectively $\forall$ and $\exists$, over individuals. Predicates are used to denote properties over individuals e.g. $P(x)$, $Q(x, y)$, \ldots As such, FOL is very expressive, and a very convenient way to formalize sentences. For instance, we can formalize the next sentence: ``all airplanes land on wheels'', with the following rule:
\begin{equation}\label{eq:alt3}
	\forall y, Airplane(y) \rightarrow Land\_on\_wheels(y)
\end{equation}
But we also know that some floatplanes are airplanes that do not land on wheels and some airplanes use skis to land on ice or snow. So, we have the following rules:
\begin{equation}\label{eq:alt35}
    \forall	y, \mathit{Ski\_airplane(y) \rightarrow Airplane(y)}
\end{equation}
\vspace*{-3mm}
\begin{equation}\label{eq:alt4}
    \forall	y, \mathit{Ski\_airplane(y) \rightarrow \lnot Land\_on\_wheels(y)}
\end{equation}
We can see that formalizations (\ref{eq:alt3}) and (\ref{eq:alt4}) are contradictory. This is because the inference in classical logic is monotonic. This property is very important in the world of mathematics, because it allows to describe lemmas previously demonstrated. But this property cannot be applied in situation where uncertain, incomplete information or exceptions have to be considered. In such situations, we would expect that by adding new information or set of formulas to a model, the set of consequences of this model might be reduced. Since the property of monotony is: $A \vdash w$ then $A \cup B \vdash w$, the problem leads directly to the general representation of common sense reasoning. By moving to a nonmonotonic framework, we can carry out the principle of explosion and nevertheless reach a conclusion.

\subsection{Default Logic} 
Default logic is one of the best known formalization for commonsense reasoning, founded by Raymond Reiter. This kind of formalization allows to infer arguments based on partial and/or contradictory information as premises \cite{reiter1980logic}. A default theory is a pair $\Delta = (D,W)$, where $D$ is a set of defaults and $W$ is a set of formulas in FOL. A default $d$ is: $\frac{A(X):B(X)}{C(X)}$, where $A(X),B(X),C(X)$ are well-formed formulas. $A(X)$ are the \textit{prerequisites}, $B(X)$ are the \textit{justifications} and $C(X)$ are the \textit{consequences}. Where $X=(x_1, x_2, x_3,\ldots,x_n)$ is a vector of (non-quantified) free variables. Intuitively a default means,``if $A(X)$ is true, and there is no evidence that $B(X)$ might be false, then $C(X)$ can be true''. 
The use of defaults implies the generation of sets containing the consequences of these defaults. Such set are called extensions. An extension can be seen as a set of beliefs of acceptable alternatives. Formally, an extension of a default theory $\Delta$ is a smallest fixed-point $E$ for which the following property holds: If $d$ is a default of $D$, whose the prerequisite is in $E$, and the negation of its justification is not in $E$, then the consequent of $d$ is in $E$ \cite{reiter1980logic}.
\begin{definition}
		Let $\Delta = (D,W)$, an \emph{extension} $E$ of $\Delta$ is define:
\end{definition}
\begin{itemize}
	\item $E = \bigcup_{i=0}^{\infty} E_i$ with:
	\item $E_0 = W$ and,
	\item for $i > 0$; $E_{i+1} = Th(E_i) \cup \{C(X) \mid \frac{A(X):B(X)}{C(X)} \in D$, $A(X)  \in E_i, \neg B(X) \not\in E\}$
\end{itemize}
Here $Th(E_i)$ is the set of formulas derived from $E_i$. A special case concerns \emph{normal default theories}, having only defaults of the form: $\frac{A(X):C(X)}{C(X)}$. The main characteristic of such default theories is that at least one extension is always guaranteed. The original version of the definition of an extension is difficult to compute in practice, since the condition $\neg B \not\in E$ assumes that $E$ is known, while $E$ is not yet calculated. In the case of normal defaults, we simply check that $E$ is an extension of $\Delta$ by replacing $\mathit{\neg B(X)\not\in E}$ by $\mathit{\neg C(X) \not\in E_i}$.
Regarding the rules (\ref{eq:alt3}) and (\ref{eq:alt4}), we can generalize the sentence \emph{``all airplanes land on wheels''} by ``\emph{generally, airplanes land on wheels}''. Having a default theory that is composed of $D = \{\frac{\mathit{Airplanes(y)} : \mathit{Land\_on\_wheels(y)}}{\mathit{Land\_on\_wheels(y)}}\}$, and a knowledge about airplanes: 
$W = \{\mathit{Floatplane(y) \rightarrow Airplane(y)}, \mathit{Floatplane(y) \rightarrow \lnot \mathit{Land\_on\_wheels(y)}}\}$. Using $D$, we can note that the prerequisite $\mathit{Airplane(y)}$ is true and the justification $\mathit{Land\_on\_wheels(y)}$ is inconsistent with $W$, because of $\mathit{Floatplane(y) \rightarrow \lnot \mathit{Land\_on\_wheels(y)}}$, then we can not conclude that floatplanes land on wheels.
But we know that some floatplanes have wheels, formally, $\mathit{W \cup \{Floatplane\_wheels(y) \rightarrow Airplane(y)\}}$. With this a new information, the prerequisite of $D$ is true and the justification is consistent, then we can conclude that there are floatplanes that have wheels and land on wheels.


\section{Embedded Computer} 
\label{sec:implementation}
\begin{figure}[ht]
	\centering
		\includegraphics[width=0.5\linewidth]{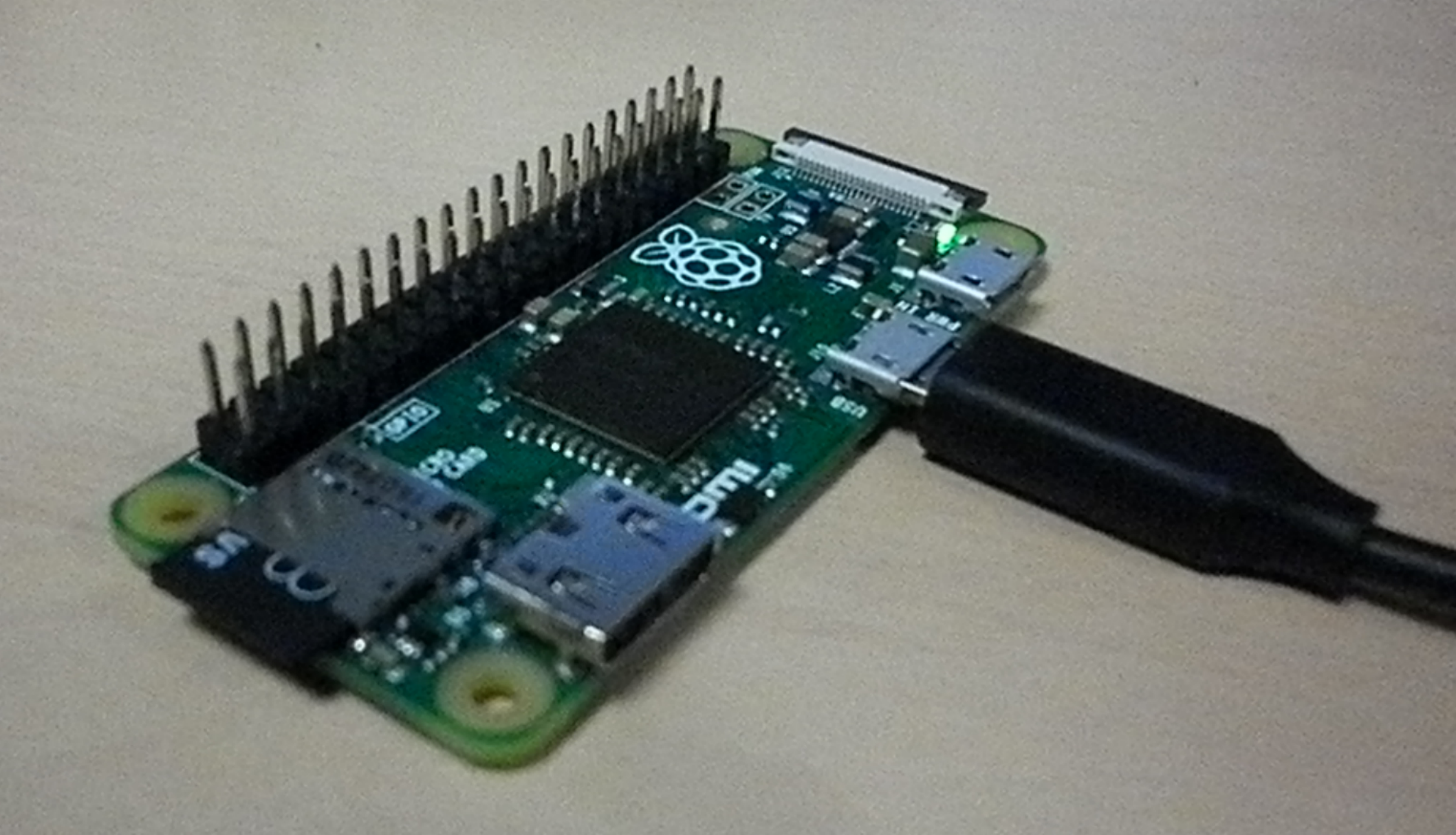}
	\caption{Embedded computer running Linux Debian}\label{fig:microcomputermcu}
\end{figure}

We use an embedded computer which is based on an ARM processor (Figure \ref{fig:microcomputermcu}, embedded computer running Linux Debian). This microcomputer supports three serial protocols communications as SPI, I2C and UART, 40 digital input/output pins\ldots Different sensors are connected to the microcomputer: a gyroscope which measures the rotations, an accelerometer that measures static and dynamic forces. Eventually, a magnetometer allows to estimate the direction by detecting the magnetic flux on earth. These three sensors can give all the information linked to a real cockpit.
The embedded computer has an operating system (OS) based on Linux Debian. Running at 1 GHz single core CPU, 512 MB of RAM and an energy consumption of 0.8 Watts. Plus, \texttt{SWI-prolog version 7.7.18} (32 bits) was installed into it.
An algorithm for calculating extensions of a default theory was implemented (Algorithm \ref{algo:defaulog}). This algorithm is coded in Prolog language and the compilation is done thanks to SWI-Prolog installed.

\begin{algorithm}[h]
	\caption{Calculation of extensions}\label{algo:defaulog}
	\begin{algorithmic}
		\Require $E = \emptyset\ $
		\Procedure{Extension}{$E = \displaystyle\cup_{i=0}^{N} E_i$}
		\State {Initialization}
		\State {Computation_of_Extension(E):}
		\While{there is a default $\frac{A(X):C(X)}{C(X)}$ that has not yet been inspected}
    	    \State {Select the default D}
    	    \State {Verify A(X) holds}
    		\State {Verify C(X) is consistent with W}
    		\State {Add C(X) to W}
    	\EndWhile
		\State {Backtracking(deleting C(X) added to W)} 
		\State {Computation_of_Extension(E)}
		\EndProcedure
	\end{algorithmic}
\end{algorithm}

\subsection{Routine of Computation}
Facts and rules follow a particular syntax. This syntax represents a normal default $d = \frac{A:C}{C}$ with a weighting: $cl(\texttt{text},\delta,A,C,\omega)$. Where \texttt{text} could be a comment describing the clause, $\delta$ could be a real fact ``\texttt{hrd}'' or default ``\texttt{def}'', $A$ and $C$ are the prerequisite and consequent respectively, and finally $\omega$ is a weighing as a priority. 
Real facts with no weighting are facts sampled at some moment and they are represented as following:
\begin{lstlisting}[language=Prolog, frame=single, breaklines=true]
	cl(``text'',hrd,[], glider(airspeed_low),[]).
	cl(``text'',hrd,[], glider(pitch_stable),[]).
	cl(``text'',hrd,[], glider(roll_stable),[]).
	cl(``text'',hrd,[], glider(altimeter_low),[]).
	cl(``text'',hrd,[], glider(variometer_zero),[]).
\end{lstlisting}
These facts are describing the states of an airplane, in our case a glider, having no inclination nor rotation, low airspeed and low altitude, and no vertical speed. We can assume here that the glider has no motion because of the states of the airplane. Due to limited space to write, we will take the following notation; $g$ for glider, $var\_zero$ for zero vertical speed, $pch\_stb$ for no inclination, $rll\_stb$ for no rotation, and $auth$ for control tower authorization. 
In the same sense, defaults are represented as following:
\begin{lstlisting}[language=Prolog, frame=single, breaklines=true, basicstyle=\footnotesize]
	cl('',def,[g(var_zero),g(pch_stb),g(rll_stb),auth],pilot(motor),[]).
\end{lstlisting}
Consider a default representing $\frac{(g(var\_zero) \land g(pch\_stb) \land g(rll\_stb) \land auth):pilot(motor)}{pilot(motor)}$, where the prerequisite is the information from the cockpit and authorization. In this example, if the context makes the default true and there are no contradictions with the conclusion, we jump to the conclusion: $pilot(motor)$\footnote{Due to the lack of space, we do not describe decision making combined with the weighting in this example.}. This is intuitively the way our Prolog program finds the solutions. Our system manages 113 rules which are as follows:

\begin{figure}[ht]
	\begin{lstlisting}[language=Prolog, frame=single, breaklines=true, basicstyle=\footnotesize]
	1 p=0,0 hrd : -> glider(airspeed_low)
	2 p=0,0 hrd : -> glider(roll_stable)
	3 p=0,0 hrd : -> glider(altimeter_low)
	4 p=0,0 hrd : -> glider(variometer_low)
	5 p=0,0 hrd : -> glider(land)
	6 p=0,0 def : glider(airspeed_zero), glider(altimeter_zero), -> glider(rest)
	7 p=0,0 def : glider(airspedd_zero), glider(-motor), -> glider(rest)
	...
	63 p=0,0 def : glider(landing), -> glider(rest_p)
	64 p=0,0 def : glider(rest_p), -> pilot(yoke_p_n)
	...
	110 p=0,0 hrd : pilot(yoke_p_n), -> -pilot(yoke_pull)
	111 p=0,0 hrd : pilot(yoke_pull), -> -pilot(yoke_p_n)
	112 p=0,0 hrd : pilot(yoke_pull), -> -pilot(yoke_push)
	113 p=0,0 hrd : pilot(yoke_push), -> -pilot(yoke_pull)
	\end{lstlisting}
	\caption{Our system with 113 rules of piloting.}
\end{figure}

\subsection{Example and Results}
Considering the facts $G$, we consult our Prolog program to know if in this context, we could take-off\ldots
\begin{equation*}
	G : g(pitch\_stable), g(roll\_stable), g(motor\_off),g(low\_alt), g(low\_airspeed)
\end{equation*}
From Algorithm \ref{algo:defaulog} programmed in Prolog, 5 different extensions are obtained. Table \ref{tab:tabcolsep2} summarizes the computed extensions with the defaults involved in each extension.
\begin{table}[h]
	\centering
	\begin{tabular}{ccccccc}
		\hline
		Extensions	& $d_{16}$ & $d_{17}$ & $d_{18}$ & $d_{19}$ & $d_{20}$ & $d_{21}$\\
		\hline
		$E_0$	& \checked& \checked& & &\checked&\\
		$E_1$	& \checked& \checked& & &&\checked\\
		$E_2$	& \checked& & &\checked &&\checked\\
		$E_3$	& &\checked & \checked& &\checked&\\
		$E_4$	& & \checked&\checked & &&\checked\\
		\hline
	\end{tabular}
	\caption{Extensions and defaults calculated}\label{tab:tabcolsep2}
\end{table}
The formalization of defaults are shown in Table \ref{tab:extension2}.

\begin{table}[h]
    \centering
    \begin{tabular}{l l}
            \vspace*{1.5mm}
	        $d_{16} = \dfrac{g(roll\_stable):pilot(yoke\_roll\_neutral)}{pilot(yoke\_roll\_neutral)}$
	        &
			$d_{17} = \dfrac{g(pitch\_stable):pilot(yoke\_pull)}{pilot(yoke\_pull)}$\\ \vspace*{1.5mm}
			$d_{18} = \dfrac{g(low\_alt):pilot(yoke\_roll\_neutral)}{pilot(yoke\_roll\_neutral)}$
		&
	        $d_{19} = \dfrac{g(low\_alt):pilot(yoke\_push)}{pilot(yoke\_push)}$\\ 
			$d_{20} = \dfrac{g(low\_alt):pilot(motor)}{pilot(motor)}$
			&
			$d_{21} = \dfrac{g(low\_alt):\lnot pilot(motor)}{\lnot pilot(motor)}$\\
    \end{tabular}
    \caption{Extensions calculated for this example.}
    \label{tab:extension2}
\end{table}

From $G$ the best extension we can choose is $E_3$ because it has the good combinations of the actions 
$\{pilot(yoke\_pull),$ $pilot(yoke\_roll\_neutral),pilot(motor)\}$ to reach the goal: {\textbf{take-off}}.
Since the extension $E_0$ has: $\{pilot(yoke\_roll\_neutral),pilot(yoke\_pitch\_neutral),pilot(motor)\}$, the result is a straight flight that is not the goal, it would probably be a solution to be able to have more speed and later take-off. The extension $E_1$ has: $\{pilot(yoke\_roll\_neutral),$ $pilot(yoke\_pitch\_neutral), \lnot pilot(motor)\}$, there is no movement on the glider because the engine is off. 

The $E_2$ extension has: $\{pilot(yoke\_roll\_neutral), pilot(yoke\_push), \lnot pilot(motor)\}$, same result as the previous extension, no movement because the engine is off. And finally, the $E_4$ extension has: $\{pilot(yoke\_pull), pilot(yoke\_pitch\_neutral), \lnot pilot(motor)\}$, same result as the previous extension, no movement because the engine is off.
To study the case where there are the maximum number of solutions, we change the facts $G$ and we obtain 13 extensions, Figure \ref{fig:newext}, with the same 113 rules. As additional information the computation of the extensions is taking 1.514 seconds.
\begin{figure}[ht]
	\begin{lstlisting}[language=Prolog, frame=single, breaklines=true, basicstyle=\footnotesize]
	EXTENSION : TRUE LITERALS : glider(airspeed_low), glider(altimeter_low), glider(descend), glider(final_approach), glider(final_approach_p), glider(land), glider(landing), glider(rest), glider(rest_p), glider(roll_stable), glider(variometer_down), pilot(yoke_p_n), pilot(yoke_r_n), pilot(-motor)
	DEFAULTS USED : [66,65,64,57,33,31,30,27,9]
	...
	EXTENSION : TRUE LITERALS : glider(airspeed_low), glider(altimeter_low), glider(land), glider(landing), glider(roll_stable), glider(takeoff), glider(takeoff_p), glider(variometer_down), pilot(motor), pilot(yoke_p_n), pilot(yoke_r_n)
	DEFAULT USED : [70,69,68,36,31,13]
	...
	EXTENSION : TRUE LITERALS : glider(airspeed_low), glider(altimeter_low), glider(climb_p), glider(land), glider(landing), glider(landing_p), glider(roll_stable), glider(takeoff), glider(variometer_down), pilot(motor_2), pilot(yoke_pull), pilot(yoke_r_n)
	DEFAULT USED : [90,72,71,62,37,31,13]
	\end{lstlisting}
	\caption{Computation of extensions: 113 instanced clauses, 5 elementary facts, and 13 extensions. Computation in 168,334 inferences, 1.478 CPU in 1.514 seconds (98\% CPU, 113881 Lips).}\label{fig:newext}
\end{figure}
As the facts change, the calculation time of the extensions, Table \ref{tab:tabcolsep}, will be variable. That is, if we have few facts, Prolog will make more logical inferences per second (Lips) to prove the consistency of the rules. However, if we have more facts, Prolog will make fewer inferences (Lips) to prove the rules.
\begin{table}[h]
	\centering
		\begin{tabular}{cccccc}
			\hline
			Facts	& Extensions	& Instanced clauses	& CPU  & Lips \\
			\hline
			7	& 13	& 115	& 95\% &114,131 \\
			5	& 13	& 113 & 98\% & 117,176 \\
			4	& 10	& 112 & 97\% &  130,098\\
			\hline
		\end{tabular}
		\caption{Comparative table on the results obtained from three different situations.\label{tab:tabcolsep}}
\end{table}
\section{Conclusion} 
\label{sec:conclusion}
We successfully installed ``\texttt{SWI-prolog version 7.7.18}'' in an embedded computer. Also nonmonotonic reasoning for piloting an airplane was implemented. Our embedded system is composed of sensors such as gyroscope, accelerometer, magnetometer, pressure sensor,\ldots Data sensors are transformed into normal defaults in order to respect a specific syntax. We tackled the problem of incomplete and uncertain information by formalizing the rules of piloting using default logic. We get good results in terms of calculation time, thanks to the use of Horn clauses and normal defaults (Considering the restriction of the embedded computer such as the low energy consumption (0.8 Watts), running at 1 GHz ARM11 (single core) and 512 Mb of RAM). Eventually, we described an example in which 5 extensions are obtained. The model implemented has 113 defaults. 
We are currently working on the decision making part in order to have a control system based on queries in Prolog. In addition we can take advantage of the infinite loops of Prolog which is one of his most outstanding tools. This allows to compute the extensions all the time while embedded computer is on.

\bibliographystyle{eptcs}
\bibliography{example}

\begin{thebibliography}{10}
\providecommand{\bibitemdeclare}[2]{}
\providecommand{\surnamestart}{}
\providecommand{\surnameend}{}
\providecommand{\urlprefix}{Available at }
\providecommand{\url}[1]{\texttt{#1}}
\providecommand{\href}[2]{\texttt{#2}}
\providecommand{\urlalt}[2]{\href{#1}{#2}}
\providecommand{\doi}[1]{doi:\urlalt{http://dx.doi.org/#1}{#1}}
\providecommand{\bibinfo}[2]{#2}

\bibitemdeclare{article}{DelScha}
\bibitem{DelScha}
\bibinfo{author}{J.~\surnamestart Delgrande\surnameend} \&
  \bibinfo{author}{T.~\surnamestart Schaub\surnameend} (\bibinfo{year}{2003}):
  \emph{\bibinfo{title}{On the relation between Reiter's default logic and its
  (major) variants.}}
\newblock {\sl \bibinfo{journal}{Seventh European Conference on Symbolic and
  Quantitative Approaches to Reasoning with Uncertainty (ECSQARU)}},
  \doi{10.1007\%2F978-3-540-45062-7_37}.

\bibitemdeclare{article}{DegrandeSchaub}
\bibitem{DegrandeSchaub}
\bibinfo{author}{J.~\surnamestart Delgrande\surnameend} \&
  \bibinfo{author}{T.~\surnamestart Schaub\surnameend} (\bibinfo{year}{2005}):
  \emph{\bibinfo{title}{Expressing Default Logic Variants in Default Logic.}}
\newblock {\sl \bibinfo{journal}{Journal of Logic and Computation}},
  \doi{10.1093/logcom/exi021}.

\bibitemdeclare{book}{kifer2018declarative}
\bibitem{kifer2018declarative}
\bibinfo{author}{Michael \surnamestart Kifer\surnameend} \&
  \bibinfo{author}{Yanhong~Annie \surnamestart Liu\surnameend}
  (\bibinfo{year}{2018}): \bibinfo{publisher}{Morgan \& Claypool},
  \doi{10.1145/3191315}.

\bibitemdeclare{inproceedings}{konolige1987relation}
\bibitem{konolige1987relation}
\bibinfo{author}{Kurt \surnamestart Konolige\surnameend}
  (\bibinfo{year}{1988}): \emph{\bibinfo{title}{On the relation between default
  and autoepistemic logic.}}
\newblock In: {\sl \bibinfo{booktitle}{Artificial Intelligence}},
  \bibinfo{volume}{35}, pp. \bibinfo{pages}{343--382},
  \doi{10.1016/0004-3702(88)90021-5}.

\bibitemdeclare{article}{McCarthya}
\bibitem{McCarthya}
\bibinfo{author}{J.~\surnamestart McCarthy\surnameend} (\bibinfo{year}{1980}):
  \emph{\bibinfo{title}{Circumscription - A form of non-monotonic reasoning.}}
\newblock {\sl \bibinfo{journal}{Artificial intelligence}}
  \bibinfo{volume}{13:1-2}, \doi{10.1016/0004-3702(80)90011-9}.

\bibitemdeclare{article}{McCarthyb}
\bibitem{McCarthyb}
\bibinfo{author}{J.~\surnamestart McCarthy\surnameend} (\bibinfo{year}{1986}):
  \emph{\bibinfo{title}{Applications of circumscription to formalizing
  common-sense knowledge.}}
\newblock {\sl \bibinfo{journal}{Artificial intelligence}}
  \bibinfo{volume}{28-1}, \doi{10.1016/0004-3702(86)90032-9}.

\bibitemdeclare{article}{reiter1980logic}
\bibitem{reiter1980logic}
\bibinfo{author}{Raymond \surnamestart Reiter\surnameend}
  (\bibinfo{year}{1980}): \emph{\bibinfo{title}{A logic for default
  reasoning}}.
\newblock {\sl \bibinfo{journal}{Artificial intelligence}}
  \bibinfo{volume}{13:1-2}, \doi{10.1016/0004-3702(80)90014-4}.

\bibitemdeclare{incollection}{reiter1981closed}
\bibitem{reiter1981closed}
\bibinfo{author}{Raymond \surnamestart Reiter\surnameend}
  (\bibinfo{year}{1981}): \emph{\bibinfo{title}{On closed world data bases}}.
\newblock In: {\sl \bibinfo{booktitle}{Readings in artificial intelligence}},
  \bibinfo{publisher}{Springer}, pp. \bibinfo{pages}{119--140},
  \doi{10.1007\%2F978-1-4684-3384-5_3}.

\bibitemdeclare{book}{russell2016artificial}
\bibitem{russell2016artificial}
\bibinfo{author}{Stuart~J \surnamestart Russell\surnameend} \&
  \bibinfo{author}{Peter \surnamestart Norvig\surnameend}
  (\bibinfo{year}{2016}): \bibinfo{publisher}{Malaysia; Pearson Education
  Limited,}, \doi{10.1145/201977.201989}.

\bibitemdeclare{book}{Sombe}
\bibitem{Sombe}
\bibinfo{author}{L.~\surnamestart Somb{\'e}\surnameend} (\bibinfo{year}{1990}):
  \bibinfo{publisher}{John Wiley \& Sons Inc}, \doi{10.1002/int.4550050403}.

\bibitemdeclare{book}{sterling1990practice}
\bibitem{sterling1990practice}
\bibinfo{author}{Leon \surnamestart Sterling\surnameend}
  (\bibinfo{year}{1990}): \bibinfo{publisher}{MIT press},
  \doi{10.1017/S026357470001599X}.

\end{thebibliography}
\end{document}